\documentclass{article}
\usepackage{spconf}
\usepackage{amsmath,amssymb}
\usepackage{graphicx}
\usepackage{etoolbox}
\setkeys{Gin}{draft=false}
\usepackage{booktabs}
\usepackage{multirow}
\usepackage{array}
\usepackage{microtype}
\usepackage{cite}
\usepackage{float}
\usepackage{xcolor}
\usepackage{flushend}

\newcommand{\method}{ASTRA-SR}

\makeatletter
\newcommand{\includegroupedpdf}[2][]{%
\begingroup
\patchcmd{\Gread@pdf}{\Gread@transgrouptrue}{\Gread@transgroupfalse}{}{}%
\includegraphics[#1]{#2}%
\endgroup}
\def\fnum@table{{\bf Tab.\ \thetable}}
\makeatother
\renewcommand{\paragraph}[1]{\par\smallskip\noindent\textbf{#1}\ }

\title{\method: ATMOSPHERIC SEEING AND TURBULENCE RESTORATION FOR ASTRONOMICAL IMAGE SUPER-RESOLUTION}
\name{Xining Ge}
\address{Hangzhou Dianzi University \\ \phantom{University affiliation} \\ \phantom{University affiliation}}

\begin{document}
\ninept
\raggedbottom
\setlength{\parskip}{0pt}
\setlength{\parindent}{1em}
\maketitle

\begin{abstract}
Ground-based planetary imaging suffers from atmospheric turbulence, sensor noise, and limited sampling, making restoration a joint denoising, deblurring, and super-resolution problem. We present \method{}, a blind single-frame restoration framework trained on a physics-grounded synthetic dataset. High-dynamic-range spacecraft RAW observations serve as clean sources, and paired LR inputs are synthesized using measured layer-integrated turbulence strengths, propagated moving phase screens, exposure-averaged spatially varying PSFs, and sensor noise. \method{} first estimates a noise-suppressed but blur-retaining LR image, then restores spatial structure through multiscale processing and reconstructs HR detail with serial spatial-amplitude refinement. It yields a 0.49 dB foreground PSNR gain over the strongest baseline approaches.
\end{abstract}

\section{Introduction}
Planetary imaging reveals atmospheric bands, storms, rings, impact structures, and surface morphology, supporting scientific monitoring and coordinated professional--amateur observations~\cite{mousis2014proam,kardasis2016proam}. Professional observatories and space missions provide high-quality measurements, but access to major facilities is scarce and cannot support routine individual use. Affordable telescopes and cameras are more accessible, although their images contain substantially less detail than spaceborne or large-observatory data. Reliable enhancement of a single observation is therefore valuable when adaptive optics, repeated lucky-imaging acquisition, or extensive multi-frame processing is unavailable.

\begin{figure*}[t]
\centering
\includegroupedpdf[width=0.965\textwidth]{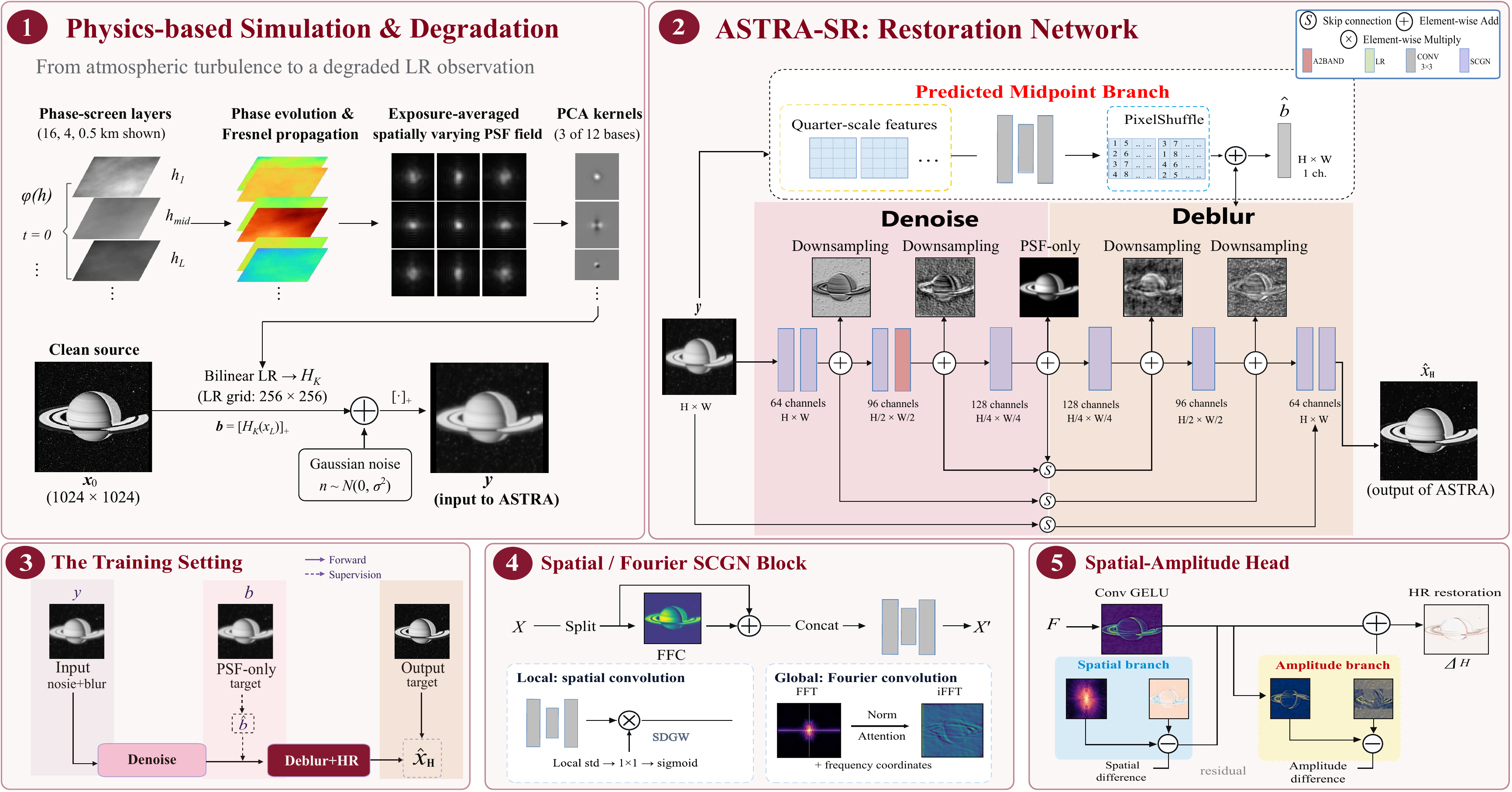}
\caption{\method{} architecture. The synthesis path generates the degraded LR input and training-only targets; the restoration path contains the three-scale SCGN backbone, intermediate LR estimation branch, stage-specific refinement, and sequential spatial-amplitude HR head.}
\label{fig:arch}
\end{figure*}

Ground-based planetary images are degraded by turbulence-induced local displacement, spatially varying blur, contrast attenuation, and temporal intensity variation. Finite exposure averages these effects into an effective point-spread function (PSF), while detector noise and limited sampling cause further information loss. Consequently, planetary bands become oversmoothed and ring edges or crater boundaries become indistinct. Recovering a high-resolution (HR) image from one low-resolution (LR) observation therefore couples denoising, deblurring, and super-resolution while risking noise amplification and artifacts~\cite{richardson1972,lucy1974,swanson2025super}.

Learning-based natural-image restoration benefits from large collections of aligned clean targets and degraded inputs~\cite{chen2022nafnet,zamir2022restormer}. Comparable astronomical pairs are rarely available because the same celestial scene cannot be captured under identical turbulent and turbulence-free conditions. Existing datasets therefore either pair observations from different instruments or synthesize turbulence from clean images. Cross-instrument pairs remain affected by differences in pixel scale, spectral response, optical PSF, detector noise, exposure, acquisition time, and viewing geometry after registration. Synthetic approaches scale more readily, but rendered RGB imagery and generic turbulence models may not represent high-dynamic-range spacecraft RAW data, measured layered turbulence, finite-exposure averaging, and instrument-specific sensor noise.

\method{} addresses this gap with a physics-grounded planetary RAW dataset and a blind single-frame restoration framework. Spacecraft observations provide high-dynamic-range, turbulence-free clean sources; measured layer-integrated turbulence strengths, propagated moving phase screens, exposure-averaged spatially varying PSFs, and sensor noise produce paired LR inputs. The network first estimates a noise-suppressed but blur-retaining LR image, then restores spatial structure through blur-aware multiscale processing, and finally reconstructs HR detail through serial spatial-amplitude refinement. Our contributions are (i) a planetary dataset of spacecraft RAW images and physically simulated LR pairs, (ii) a staged spatial-frequency framework for denoising and deblurring, and (iii) a serial spatial-amplitude HR head.

\section{Related Work}

\label{sec:related}

General image restoration uses convolutional priors~\cite{chen2022nafnet}, Transformer and frequency-domain modeling~\cite{zamir2022restormer,kong2023fftformer}, state-space architectures~\cite{guo2024mambair,guo2025mambairv2}, and diffusion priors~\cite{yue2023resshift,lin2024diffbir,wu2024seesr}. Recent work also explores data-centric training, self-ensemble, complementary branch fusion, and efficient super-resolution~\cite{chang2026datacentric,chang2026ensemble,ge2026dualbranch,ren2026ntire}. Explicit inverse formulations include plug-and-play restoration~\cite{zhang2022dpir}, blind deconvolution~\cite{ren2020selfdeblur,chung2023blinddps}, and distribution matching~\cite{meanti2025ddm}.

Astronomical restoration additionally depends on atmospheric propagation and instrument-specific acquisition. P2S accelerates turbulence synthesis~\cite{mao2021p2s}; PlaNet combines synthetic planetary turbulence with variable-frame restoration~\cite{xia2025planet}; FluxFlow uses co-registered ground-to-space pairs~\cite{liu2026fluxflow}; and recent methods model adaptive-optics super-resolution or video turbulence~\cite{swanson2025super,zhang2025learning}. Sensor-aware RAW processing, physical-degradation benchmarks, and noise modeling~\cite{liu2026rawild,liu2026realx3d,cao2023noiseflow,liu2026deepsky} complement self-supervised scientific denoising~\cite{lequyer2022noise2fast,li2023srdtrans,liu2025tdr,guo2026asteris}. \method{} focuses on blind single-frame planetary RAW restoration under spatially varying, exposure-averaged turbulence and sensor noise.

\section{Physics-grounded synthesis}
\label{sec:synthesis}
We define a forward degradation model that synthesizes degraded LR observations from clean planetary RAW images. The left portion of Fig.~\ref{fig:arch} illustrates the synthesis pipeline from atmospheric propagation and sensor corruption to paired LR observations and HR targets.

\subsection{Atmospheric Synthesis and PSF Formation}
\label{sec:atmospheric-synthesis}
The simulator randomly samples six-layer turbulence strengths at $\{0.5,1,2,4,8,16\}$ km from real MASS observation records\footnote{The layered measurements are taken from the European Southern Observatory (ESO) Paranal Astronomical Site Monitor MASS database.}~\cite{esoasm}. The stored values are layer-integrated strengths $J_{\ell}=\int C_n^2(h)\,dh$ in $10^{-15}\,\mathrm{m}^{1/3}$ rather than pointwise samples. For $k_0=2\pi/\lambda$,
\begin{equation}
J_{\ell}=\int C_n^2(h)\,dh,\qquad
r_0^{-5/3}=0.423k_0^2\sum_{\ell}J_{\ell}.
\end{equation}

\begin{figure}[!tp]
\centering
\includegroupedpdf[width=\columnwidth]{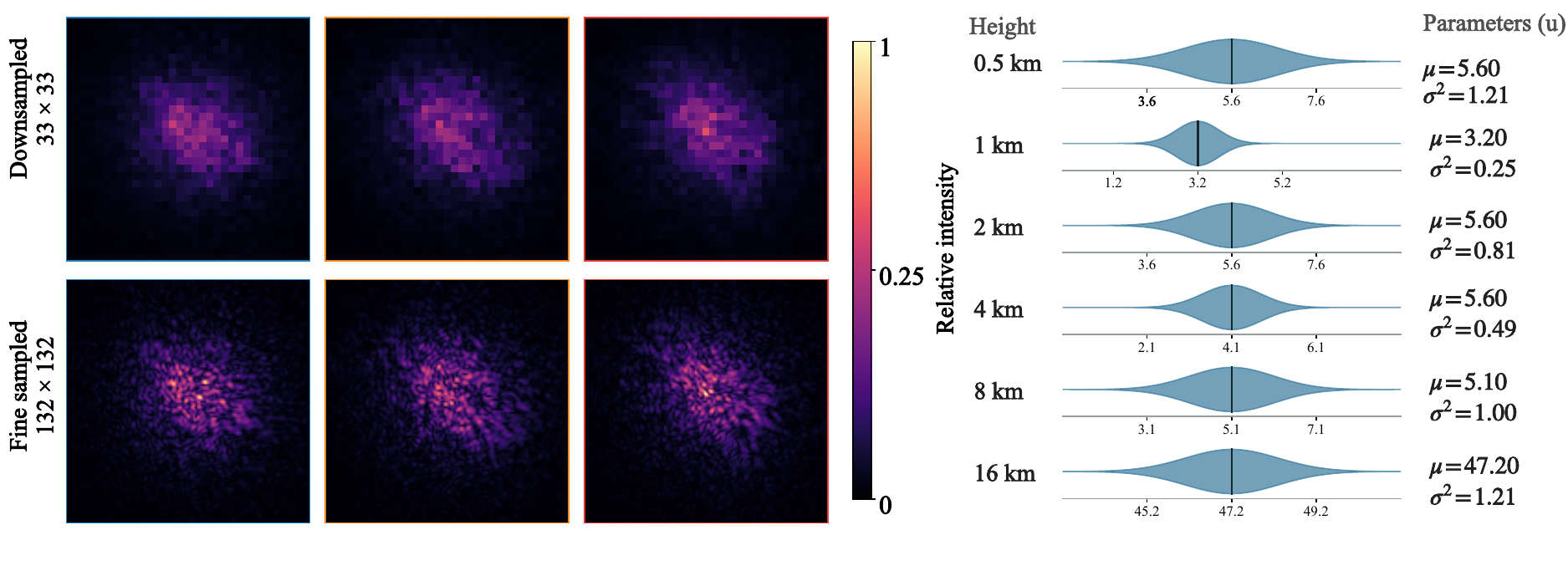}
\caption{Representative PSFs and layer-strength sampling models for an illustrated MASS profile.}
\label{fig:psf}
\end{figure}

Each atmospheric layer is modeled by a wind-driven phase screen under the frozen-flow assumption. The optical field is propagated successively between layers by split-step Fresnel propagation, followed by Fraunhofer propagation through the telescope pupil to obtain the focal-plane intensity~\cite{por2018hcipy}. With $M$ temporal samples,
\begin{equation}
 u_{\ell+1}^{\alpha,t}=\mathcal{P}_{\Delta h_{\ell}}\!\left(u_{\ell}^{\alpha,t}e^{i\phi_{\ell}^{t}}\right),\quad
 k_{\alpha}=\mathcal{N}\!\left(\frac{1}{M}\sum_{t=1}^{M}\left|\mathcal{F}\{Au_g^{\alpha,t}\}\right|^2\right),
\end{equation}
where $\mathcal{N}$ performs resampling, recentering, non-negative clipping, and unit-sum normalization. The propagated exposure PSFs are resampled to the LR lattice and act as effective discrete PSFs in the LR-domain forward model~\cite{xie2026physically,zeng2026continuous}.

\subsection{Planetary RAW Dataset and LR Observation Formation}
\label{sec:lr-formation}
We collect approximately 400,000 raw observations from the Cassini Imaging Science Subsystem (ISS) and manually remove corrupted, severely noisy, and near-duplicate frames. This process yields approximately 20,000 high-dynamic-range images of planets, asteroids, and natural satellites. Spaceborne capture preserves faint and bright-region detail without ground-atmosphere turbulence. Multiple degradations produce 63,582 training samples and 1,355 source-disjoint test samples.

Each clean source $x_0$ is independently resampled to form the LR image $x_L=R_L(x_0)$ and HR target $x_H=R_H(x_0)$. Blur is applied on the LR grid. A mean kernel $\bar{k}$ and 12 PCA basis kernels $\{k_j\}$ represent the spatially varying PSF field,
\begin{equation}
\begin{aligned}
 b(p)&=\left[(\bar{k}*x_L)(p)+\sum_{j=1}^{12}c_j(p)(k_j*x_L)(p)\right]_{+},\\
 y&=[b+n]_{+},\qquad n\sim\mathcal{N}(0,\sigma^2).
\end{aligned}
\end{equation}
Inference receives only $y$. The PSF, turbulence, noise, and auxiliary-target information are never provided as inputs.

\section{\method{}}
\label{sec:method}
\begin{figure}[!tp]
\centering
\includegraphics[width=0.96\columnwidth]{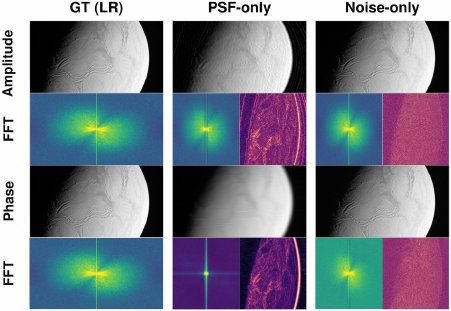}
\caption{Fourier-component restoration analysis under full, PSF-only, and noise-only degradation.}
\label{fig:fourier-restoration}
\end{figure}

The right portion of Fig.~\ref{fig:arch} presents the three stages of \method{}. Intermediate LR estimation suppresses noise while retaining blur information, blur-aware LR restoration recovers spatial and spectral structure, and sequential spatial-amplitude reconstruction produces the final HR image. To guide the placement of frequency processing, we decompose synthesized observations into Fourier magnitude and phase and evaluate recovery when each component is restored toward its clean counterpart. The diagnosis in Fig.~\ref{fig:fourier-restoration} shows that magnitude restoration yields larger gains than phase restoration under full and PSF-only degradation, motivating frequency-aware LR processing and amplitude-aware HR refinement~\cite{cui2025adair}.

\subsection{Intermediate LR Estimation}
\label{sec:intermediate-estimation}
Given the degraded LR observation $y$, a shallow head extracts features that are processed by SCGN blocks at full, half, and quarter resolution. PixelUnshuffle and convolution connect adjacent scales. At half resolution, a frequency-band adapter routes low-, mid-, and high-frequency responses while preserving spatial information before the quarter-scale bottleneck. The resulting feature $Q$ predicts
\begin{equation}
\hat{b}=y+\mathrm{PS}_4\!\left(C_b(Q)\right).
\end{equation}
Because the target $b$ removes sensor noise while retaining PSF blur, this auxiliary supervision encourages early denoising without premature deconvolution. The estimate $\hat{b}$ also provides blur-aware context for the next stage.

\subsection{Blur-Aware LR Restoration}
\label{sec:lr-restoration}
The LR restoration pathway fuses multiscale features through cross-scale skip connections. Each block combines SCGN processing with dilated depthwise branches to capture broad degradation patterns and local structure. At full LR resolution, an overlapping Patch-Fourier refiner corrects residual blur in local complex spectra. A tail convolution predicts residual $r$, modulated by a bounded global gain $g$ and the local gain $\ell(y,r,\hat{b})$,
\begin{equation}
\hat{x}_L=y+g(y)\,\ell(y,r,\hat{b})\odot r.
\end{equation}
This update adapts restoration strength across images and spatial locations, while the bounded gains limit noise over-amplification.

\subsection{Sequential Spatial-Amplitude HR Reconstruction}
\label{sec:hr-reconstruction}
Let $H_0$ denote the feature entering the HR reconstruction head. A spatial branch first refines local structure, after which an amplitude branch adjusts the spectral response,
\begin{equation}
H_1=H_0+\Delta_s(H_0),\qquad H_2=H_1+\Delta_f(H_1).
\end{equation}
For $\Delta_f$, pooled local windows are transformed into the Fourier domain. Their channel-averaged log magnitudes predict a bounded gain $1+a$, which adjusts magnitude while preserving phase. The refined feature $H_2$ is mapped to an HR residual and added to bilinearly upsampled $\hat{x}_L$,
\begin{equation}
\hat{x}_H=\mathrm{Bilinear}_2(\hat{x}_L)+\mathrm{PS}_2\!\left(C_H(H_2)\right).
\end{equation}
Training jointly constrains HR intensity, image gradients, and the intermediate LR estimate,
\begin{equation}
\mathcal{L}=\|\hat{x}_H-x_H\|_1+0.1\|\nabla\hat{x}_H-\nabla x_H\|_1
+0.1\|D_4(\hat{b})-D_4(b)\|_1.
\end{equation}
The three terms supervise HR intensity, image gradients, and intermediate denoising. The frequency-band adapter and Patch-Fourier refiner therefore operate at complementary LR stages, while the sequential head performs final HR spatial-amplitude refinement.

\begin{table}[!tp]
\centering
\caption{Quantitative comparison on 1,355 synthetic test images.}
\label{tab:main}
\setlength{\tabcolsep}{2.5pt}
\fontsize{8}{9}\selectfont
\begin{tabular*}{\columnwidth}{@{\extracolsep{\fill}}lcccc@{}}
\toprule
Method & Venue & PSNR$\uparrow$ & SSIM$\uparrow$ & Obj.-PSNR$\uparrow$ \\
\midrule
Bicubic \cite{keys1981cubic} & -- & 29.868 & 0.778 & 25.547 \\
Bilinear \cite{lehmann1999survey} & -- & 29.757 & 0.776 & 25.427 \\
\midrule
NAFNet \cite{chen2022nafnet} & ECCV'22 & 35.275 & 0.843 & 31.434 \\
Restormer \cite{zamir2022restormer} & CVPR'22 & 32.615 & 0.819 & 28.602 \\
FFTformer \cite{kong2023fftformer} & CVPR'23 & 35.131 & 0.844 & 31.547 \\
SMFANet \cite{zheng2024smfanet} & ECCV'24 & 33.605 & 0.836 & 29.605 \\
PlaNet  \cite{xia2025planet} & AAAI'25 & 35.141 & 0.841 & 31.427 \\
RDBM \cite{wang2026residual} & CVPR'26 & 34.618 & 0.844 & 31.661 \\
SCGN \cite{li2026scgn} & CVPR'26 & 35.138 & 0.846 & 31.620 \\
StarIR \cite{cui2026starir} & TPAMI'26 & 35.218 & 0.844 & 31.553 \\
\midrule
\textbf{\method{}} & \textbf{Ours} & \textbf{35.824} & \textbf{0.849} & \textbf{32.148} \\
\bottomrule
\end{tabular*}
\end{table}


\begin{figure*}[!tp]
\centering
\IfFileExists{Figure4.pdf}{%
\includegraphics[width=\textwidth]{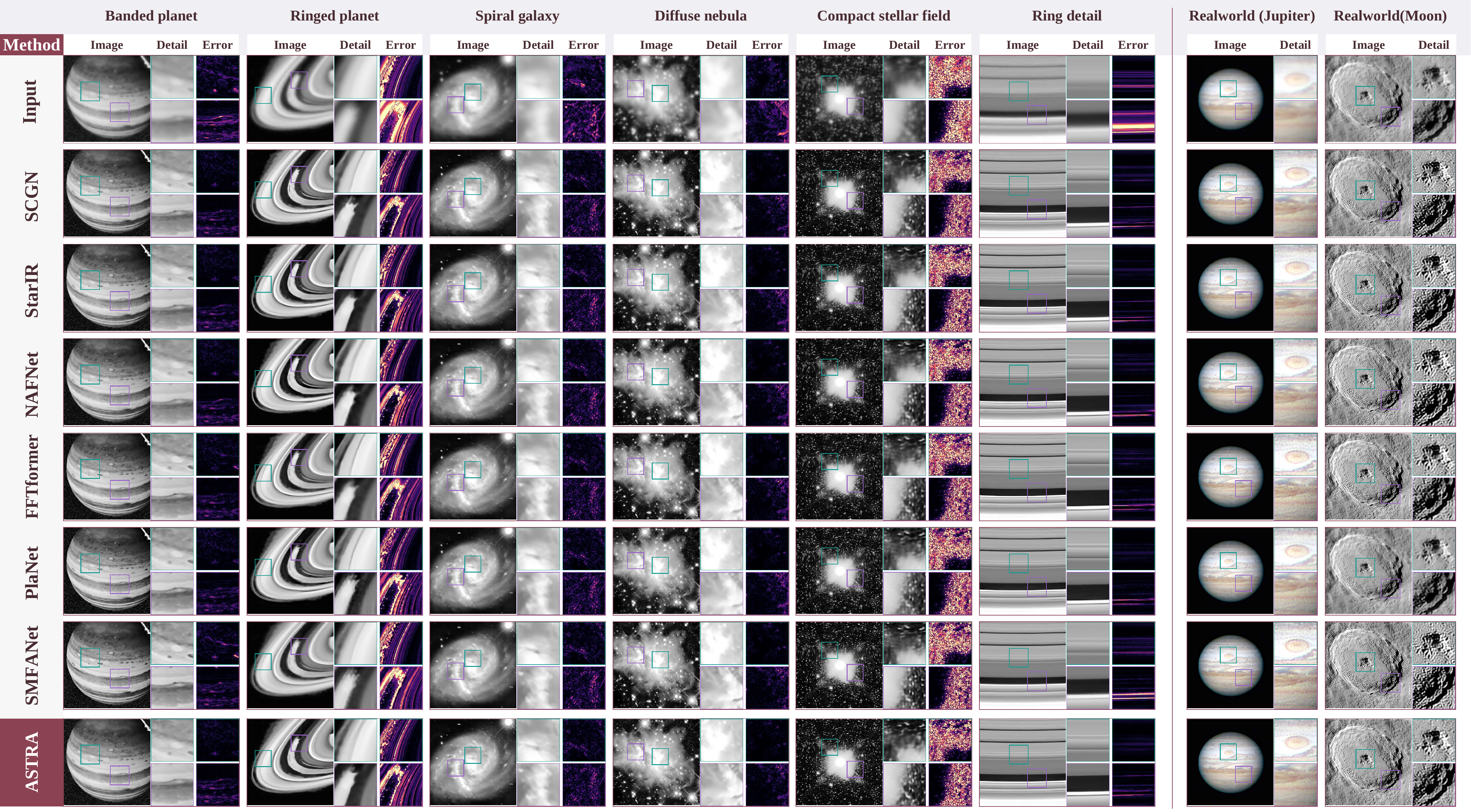}%
}{%
\fbox{\parbox[c][4cm][c]{0.94\textwidth}{\centering
\textbf{Figure file missing: \texttt{Figure4.pdf}}\\[1ex]
Upload the original figure to restore this comparison.}}%
}
\caption{Qualitative comparisons on paired synthetic data and real observations. All methods use identical coordinates and zoom regions. The real Jupiter and lunar observations have no reference targets.}
\label{fig:qual}
\end{figure*}

\section{Experiments}
\paragraph{Implementation Details and Metrics.}
Each sample maps a single-channel $256\times256$ degraded LR input to a $512\times512$ HR target. The three-scale SCGN backbone uses widths $(64,96,128)$ with block layouts $(2,1,1)$ for estimation and $(1,1,2)$ for restoration. The local branches use dilation rates $\{1,4,9\}$, the global and local residual gains are bounded to $(0.5,1.5)$, and the amplitude branch operates on pooled $32\times32$ windows with modulation scale 0.10. Training intensities are divided by 2500, giving a normalized Gaussian noise standard deviation of $8\times10^{-4}$. Predictions are clipped to the physical range and normalized to $[0,1]$. We report macro PSNR, SSIM, and Object-PSNR by removing background. Matched ablations are trained from scratch for 20 epochs under the same optimization settings.

\subsection{Quantitative Evaluation on Synthetic Dataset}
Fig.~\ref{fig:fourier-restoration} reports the quantitative Fourier-component restoration analysis used to guide the network design. Under full degradation, magnitude restoration improves PSNR from 31.164 to 34.903 dB, while phase restoration reaches 31.558 dB. The corresponding PSF-only values for the degraded input, magnitude restoration, and phase restoration are 31.195, 35.324, and 31.518 dB. Noise-only degradation benefits similarly from restoring either component.

\method{} achieves the best PSNR, SSIM, and Object-PSNR among all evaluated methods. Based on the reported table values, it improves over the strongest baseline for each metric by 0.549 dB, 0.003, and 0.487 dB, respectively. The substantial foreground-region PSNR margin further shows that the improvement extends to compact astronomical structures rather than arising only from background smoothing.

\begin{table}[!t]
\centering
\caption{Matched ASTRA ablation by module dimension.
$\Delta$ is computed from unrounded PSNR values.}
\label{tab:ablation}
\setlength{\tabcolsep}{3pt}
\renewcommand{\arraystretch}{1.03}
\fontsize{8}{8.8}\selectfont
\begin{tabular}{@{}>{\raggedright\arraybackslash}p{0.20\columnwidth}>{\raggedright\arraybackslash}p{0.47\columnwidth}cc@{}}
\toprule
Dimension & Variant & PSNR & $\Delta$ (dB) \\
\midrule
Full model & ASTRA-SR full model & \textbf{35.824} & -- \\
\midrule
\multirow{2}{*}{LR estimation} & w/o frequency-band adapter & 35.734 & 0.090 \\
& w/o intermediate LR supervision & 35.719 & 0.105 \\
\midrule
\multirow{3}{*}{LR restoration} & w/o dilated branches & 35.713 & 0.112 \\
& w/o Patch-Fourier refiner & 35.739 & 0.085 \\
& w/o local residual gain & 35.733 & 0.091 \\
\midrule
\multirow{3}{*}{HR recon.} & w/o spatial branch & 35.742 & 0.082 \\
& w/o amplitude branch & 35.411 & 0.413 \\
& w/o both branches & 34.940 & 0.885 \\
\bottomrule
\end{tabular}
\end{table}

\subsection{Qualitative Evaluation on Real-World Dataset}
Fig.~\ref{fig:qual} includes real Jupiter and lunar observations without reference targets. On Jupiter, \method{} produces clearer band boundaries and localized atmospheric structures while retaining smooth large-scale intensity variation. On the lunar observation, crater rims and fine surface transitions become more distinct without obvious periodic ringing or isolated artifacts. Although quantitative assessment is unavailable, these results provide qualitative evidence of generalization to real-world data.

On the paired synthetic cases, \method{} preserves localized detail with fewer structured residuals near object boundaries.

\subsection{Ablation Study}
Removing the amplitude branch reduces PSNR by 0.413 dB, while removing both HR branches causes a 0.885 dB drop, confirming that spatial and amplitude reconstruction are complementary. Removing intermediate LR supervision or the dilated branches decreases PSNR by 0.11 and 0.11 dB, respectively, which supports the staged separation of noise suppression and blur recovery. The frequency-band adapter, Patch-Fourier refiner, and local residual gain yield smaller but consistent gains, showing that each stage contributes without dominating the complete model.

\section{Conclusion}
\method{} combines measurement-driven atmospheric simulation with blur-aware multiscale restoration and serial spatial-amplitude HR reconstruction for blind astronomical super-resolution. Intermediate supervision, Fourier-component analysis, and matched ablations support its staged, amplitude-aware design. Because paired real-world references and astronomical restoration data remain scarce, broader quantitative evaluation across telescope configurations and observing conditions is left for future study.

\clearpage
{\fontsize{7.3}{7.8}\selectfont
\let\originalthebibliography\thebibliography
\renewcommand{\thebibliography}[1]{%
\originalthebibliography{#1}%
\setlength{\itemsep}{-0.2ex}%
\setlength{\parsep}{0pt}}
\bibliographystyle{IEEEbib}
\bibliography{refs}
}

\end{document}